\documentclass[letterpaper, 10 pt, conference]{ieeeconf}  

\IEEEoverridecommandlockouts                              

\usepackage{amsmath} 
\usepackage{amssymb}  
\usepackage{cite}
\usepackage{graphicx}

\usepackage{soul,color} 
\usepackage[ruled,norelsize,linesnumbered,lined,commentsnumbered]{algorithm2e}

\makeatletter
\let\NAT@parse\undefined
\makeatother
\usepackage{hyperref}
\usepackage{cleveref}

\makeatletter
\newcommand{\removelatexerror}{\let\@latex@error\@gobble}
\makeatother

\title{\LARGE \bf
	Integrating Deep Reinforcement Learning with Model-based Path Planners for Automated Driving  
}

\author{Ekim Yurtsever$^{*, \dagger}$, Linda Capito$^{*, \dagger}$, Keith Redmill$^{*}$ and Umit Ozguner$^{*}$
	\thanks{$^{*}$E. Yurtsever, L. Capito, K. Redmill and U. Ozguner are with The Ohio State University, Ohio, US.
		}%
	\thanks{$\dagger$ These authors contributed equally to this work}
	\thanks{Corresponding author: Ekim Yurtsever, ekimyurtsever@gmail.com
	}%
}

\begin{document}

	\maketitle
	\thispagestyle{empty}
	\pagestyle{empty}

\begin{abstract}

Automated driving in urban settings is challenging. Human participant behavior is difficult to model, and conventional, rule-based Automated Driving Systems (ADSs) tend to fail when they face unmodeled dynamics. On the other hand, the more recent, end-to-end Deep Reinforcement Learning (DRL) based model-free ADSs have shown promising results. However, pure learning-based approaches lack the hard-coded safety measures of model-based controllers. Here we propose a hybrid approach for integrating a path planning pipe into a vision based DRL framework to alleviate the shortcomings of both worlds. In summary, the DRL agent is trained to follow the path planner's waypoints as close as possible. The agent learns this policy by interacting with the environment. The reward function contains two major terms: the penalty of straying away from the path planner and the penalty of having a collision. The latter has precedence in the form of having a significantly greater numerical value. Experimental results show that the proposed method can plan its path and navigate between randomly chosen origin-destination points in CARLA, a dynamic urban simulation environment. Our code is open-source and available online \footnote{\label{footnote1}\href{https://github.com/Ekim-Yurtsever/Hybrid-DeepRL-Automated-Driving}{https://github.com/Ekim-Yurtsever/Hybrid-DeepRL-Automated-Driving}}.






\end{abstract}

\section{Introduction}

Automated Driving Systems (ADSs) promise a decisive answer to the ever-increasing transportation demands. However, widespread deployment is not on the horizon as state-of-the-art is not robust enough for urban driving. The recent Uber accident \cite{kohli2019enabling} is an unfortunate precursor: the technology is not ready yet.    

There are two common ADS design choices \cite{yurtsever2019survey}. The first one is the more conventional, model-based, modular pipeline approach \cite{urmson2008autonomous, levinson2011towards, wei2013towards, broggi2013extensive, maddern20171, akai2017autonomous, guizzo2011google, ziegler2014making}. A typical pipe starts with a perception module. Robustness of perception modules has been increased greatly due to the recent advent of deep Convolutional Neural Networks (CNN) \cite{krizhevsky2012imagenet}. The pipe usually continues with scene understanding \cite{cordts2016cityscapes}, assessment \cite{yurtsever2019risky}, planning \cite{mcnaughton2011motion} and finally ends with motor control. The major shortcomings of modular model-based planners can be summarized as complexity, error propagation, and lack of generalization outside pre-postulated model dynamics. 

The alternative end-to-end approaches \cite{chen2015deepdriving, pomerleau1989alvinn, muller2006off, bojarski2016end, xu2017end, sallab2017deep, kendall2019learning, baluja1996evolution, koutnik2013evolving, makantasis2019deep} eliminated the complexity of conventional modular systems. With the recent developments in the machine learning field, sensory inputs now can directly be mapped to an action space. Deep Reinforcement Learning (DRL) based frameworks can learn to drive from front-facing monocular camera images directly \cite{kendall2019learning}. However, the lack of hard-coded safety measures, interpretability, and direct control over path constraints limit the usefulness of these methods.

\begin{figure}
	\centering
	\includegraphics[width=1\columnwidth]{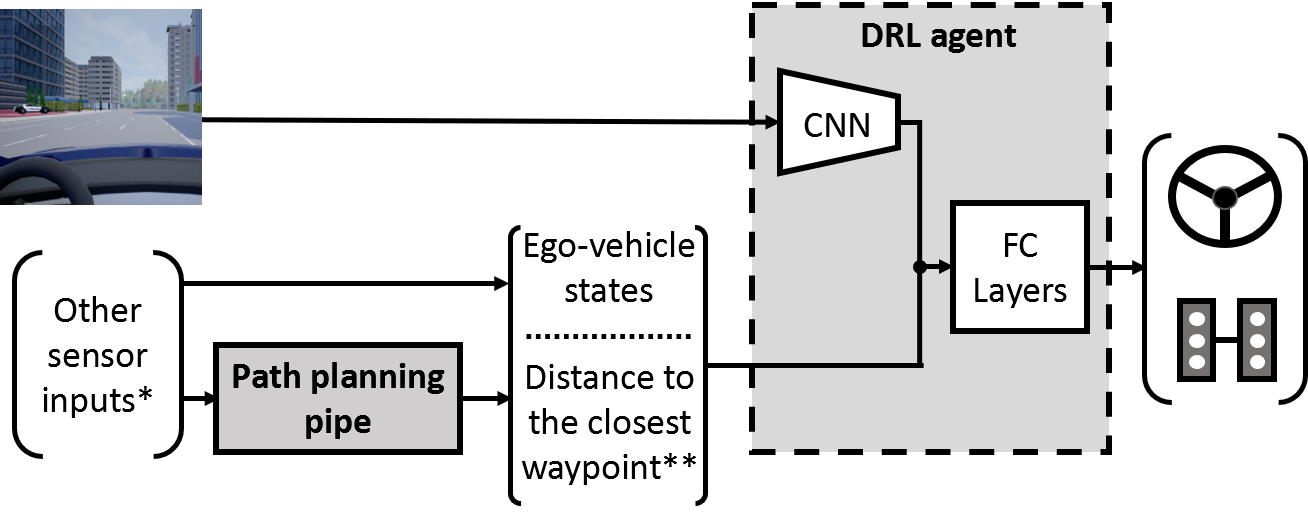}    
	\caption{An overview of our framework. FC stands for Fully Connected layers. The proposed system is a hybrid of a model-based planner and a model-free DRL agent. *Other sensor inputs can be anything the conventional pipe needs. ** We integrate planning into the DRL agent by adding `distance to the closest waypoint' into our state-space, where the path planner gives the closest waypoint. Any kind of path planner can be integrated into the DRL agent with the proposed method.} 
	\label{fig_overview}
	
\end{figure}

We propose a hybrid methodology to mitigate the drawbacks of both approaches. In summary, the proposed method integrates a short pipeline of localization and path planning modules into a DRL driving agent. The training goal is to teach the DRL agent to oversee the planner and follow it if it is safe to follow. The proposed method was implemented with a Deep Q Network (DQN) \cite{mnih2015human} based RL agent and the A* \cite{hart1968formal} path planner. First, the localization module outputs the ego-vehicle position. With a given destination point, the path planner uses the A* algorithm \cite{hart1968formal} to generate a set of waypoints. The distance to the closest waypoint, along with monocular camera images and ego-vehicle dynamics, are then fed into the DQN based RL agent to select discretized steering and acceleration actions. During training, the driving agent is penalized for making collisions and being far from the closest waypoint asymmetrically, with the former term having precedence. We believe this can make the agent prone to follow waypoints during free driving but have enough flexibility to stray from the path for collision avoidance using visual cues. An overview of the proposed approach is shown in Figure \ref{fig_overview}.

The major contributions of this work can be summarized as follows:
\begin{itemize}
	\item A general framework for integrating path planners into model-free DRL based driving agents  
	\item Implementation of the proposed method with an A* planner and a DQN RL agent. Our code is open-source and available online\textsuperscript{\ref{footnote1}}. 

\end{itemize}

The remainder of the paper is organized in five sections. A brief literature survey is given in Section \ref{sec_related}. Section \ref{sec_method} explains the proposed methodology and is followed by experimental details in Section \ref{sec_experiments}. Results are discussed in Section \ref{sec_results} and a short conclusion is given in Section \ref{sec_conclusion}. 

\section{Related Works}\label{sec_related}


End-to-end driving systems use a single algorithm/module to map sensory inputs to an action space. ALVINN \cite{pomerleau1989alvinn} was the first end-to-end driving system and utilized a shallow, fully connected neural network to map image and laser range inputs to a discretized direction space. The network was trained in a supervised fashion with labeled simulation data. More recent studies employed real-world driving data and used convolutional layers to increase performance \cite{bojarski2016end}. However, real-world urban driving has not been realized with an end-to-end system yet. 

A CNN based partial end-to-end system was introduced to map the image space to a finite set of intermediary ``affordance indicators" \cite{chen2015deepdriving}. A simple controller logic was then used to generate driving actions from these affordance indicators. Chauffer Net \cite{bansal2018chauffeurnet} is another example of a mid-to-mid system. These systems benefit from robust perception modules on the one end, and rule-based controllers with hard-coded safety measures on the other end. 

All the methods mentioned above suffer from shortcomings of supervised learning—namely, a significant dependency on labeled data, overfitting, and lack of interpretability. Deep Reinforcement Learning (DRL) based automated driving agents \cite{sallab2017deep, kendall2019learning} replaced the need for huge amounts of labeled data with online interaction. DRL agents try to learn the optimum way of driving instead of imitating a target human driver. However, the need for interaction raises a significant issue. Since failures cannot be tolerated for safety-critical applications, in almost all cases, the agent must be trained in a virtual environment. This adds the additional virtual-to-real transfer learning problem to the task. In addition, DRL still suffers from a lack of interpretability and hard-coded safety measures.

A very recent study \cite{hoel2019combining} focused on general tactical decision making for automated driving using the AlphaGo Zero algorithm \cite{silver2017mastering}. AlphaGo Zero combines tree-search with neural networks in a reinforcement learning framework, and its implementation to the automated driving domain is promising.  However, this study \cite{hoel2019combining} was limited to only high-level tactical driving actions such as staying on a lane or making a lane change. 

Against this backdrop, here we propose a hybrid DRL-based driving automation framework. The primary motivation is to integrate path-planning into DRL frameworks for achieving a more robust driving experience and a faster learning process.

\section{Proposed Method}\label{sec_method}
\begin{figure}
	\centering
	\includegraphics[width=1\columnwidth]{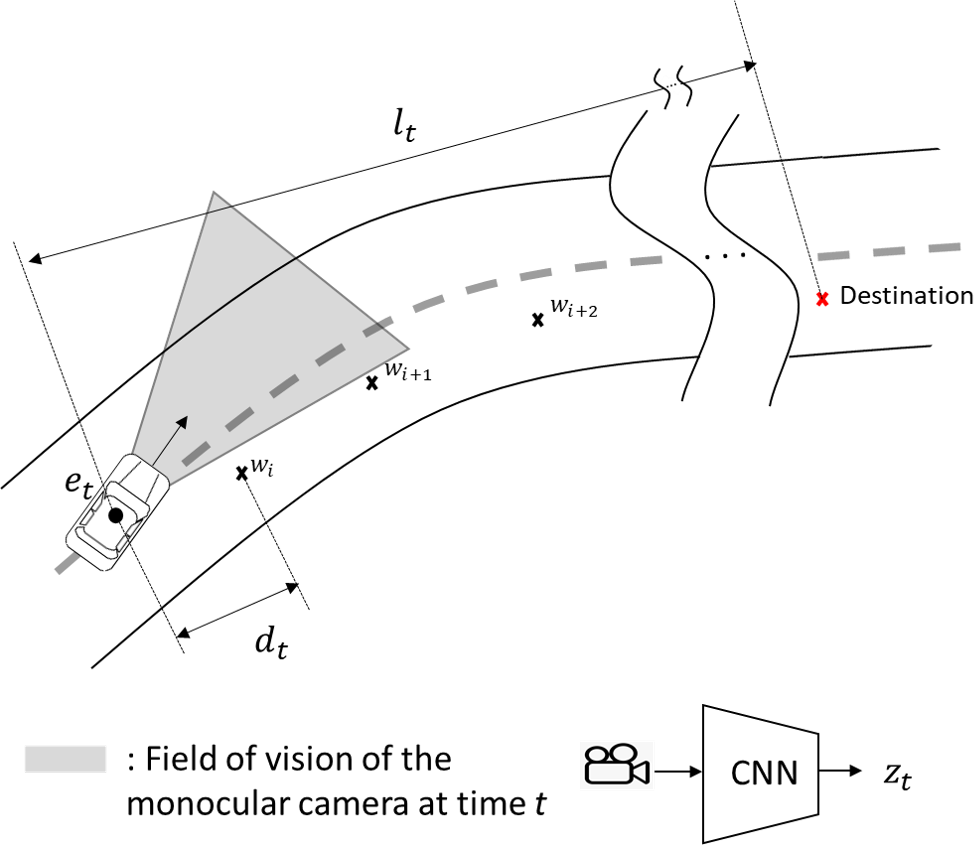}    
	\caption{Illustration of state $s_{t} \simeq (z_{t}, e_{t}, d_{t})$ and distance to the final destination $l_{t}$ at time $t$. Waypoints $w \in W$ are to be obtained from the path planner.}
	\label{fig_states}
\end{figure}

\subsection{Problem formulation}

In this study, automated driving is defined as a Markov Decision Process (MDP) with the tuple of $(S, A, P, r)$. We integrate path-planning into the MDP by adding $d$, distance to the closest waypoint, to the state-space.

\begin{figure*}
	\centering
	\includegraphics[width=2\columnwidth]{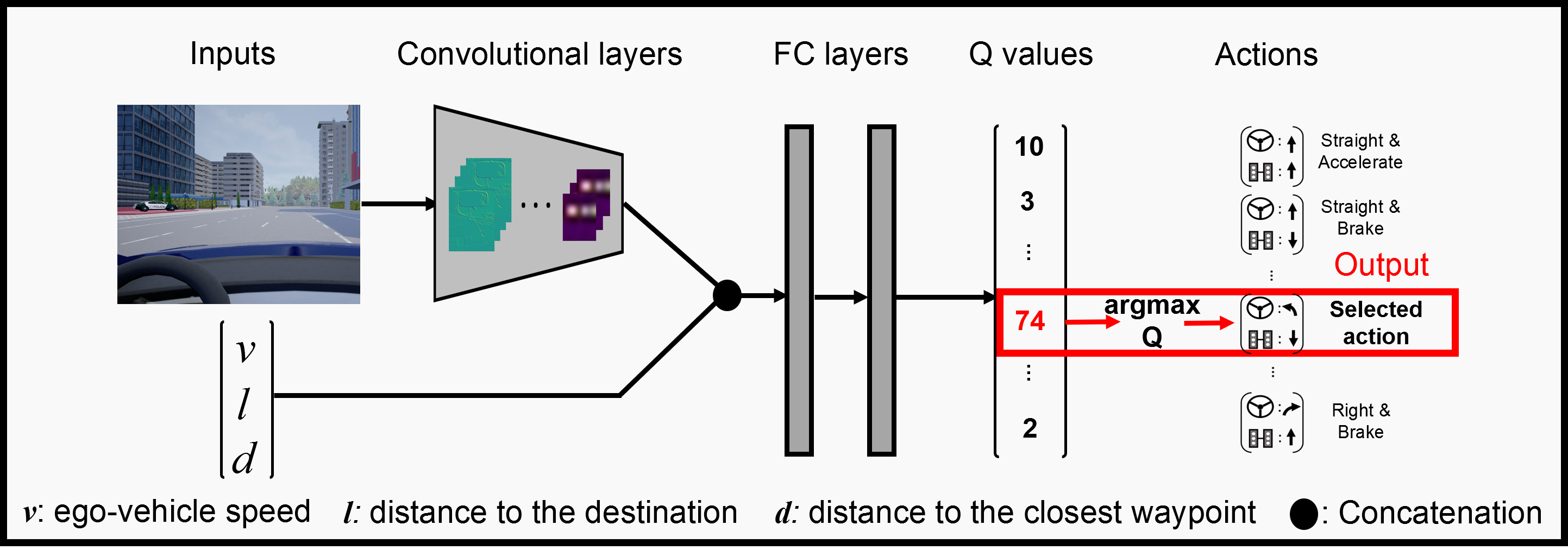}    
	\caption{The DQN based DRL agent. FC stands for fully connected. After training, the agent selects the best action by taking the argmax of predicted Q values.  }
	\label{fig_dqn}
\end{figure*}

\begin{itemize}
	\item [$S$:] A set of states. We associate observations made at time \textit{t} with the state $s_{t}$ as $s_{t} \simeq (z_{t}, e_{t}, d_{t})$ where; 1) $z_{t}=f_{cnn}(I_{t})$ is a visual feature vector which is extracted using a deep CNN from a single image $I_{t}$ captured by a front-facing monocular camera. 2) $e_{t}$ is a vector of ego-vehicle states including speed and location 3) $d_{t}$ is the distance to the closest waypoint obtained from the model-based path planner. $d_{t}$ is the key observation which links model-based path planners to the MDP. 
	\item [$A$:] A set of discrete driving actions illustrated in Figure \ref{fig_dqn}. Actions consist of discretized steering angle and acceleration values. The agent executes actions to change states.
	\item[$P$:] The transition probability $P_{t} = \textrm{Pr}(s_{t+1}|s_{t}, a_{t})$. Which is the probability of reaching state $s_{t+1}$ after executing action $a_{t}$ in state $s_{t}$.
	\item[$r$:] A reward function $r(s_{t+1}, s_{t}, a_{t})$. Which gives the instant reward of going from state $s_{t}$ to $s_{t+1}$ with $a_{t}$.     
\end{itemize}

The goal is to find a policy function $\pi(s_{t}) = a_{t}$ that will select an action given a state such that it will maximize the following expectation of cumulative future rewards where $s_{t+1}$ is taken from $P_{t}$. 

\begin{equation}
\label{eqn_classification}
\mathbb{E} \left(  \sum _{t=0}^{\infty }{\gamma ^{t}r(s_{t},s_{t+1},a_{t})} \right) 
\end{equation}

\noindent Where $\gamma$ is the discount factor, which is a scalar between $0 \leq \gamma \leq 1$ that determines the relative importance of later rewards with respect to previous rewards. We fix the horizon for this expectation with a finite value in practice.     

Our problem formulation is similar to a previous study \cite{kendall2019learning}, the critical difference being the addition of $d_{t}$ to the state space and the reward function. An illustration of our formulation is shown in Figure \ref{fig_states}.

\subsection{Reinforcement Learning}

Reinforcement learning is an umbrella term for a large number of algorithms derived for solving the Markov Decision Problems (MDP) \cite{kendall2019learning}. 

In our framework, the objective of reinforcement learning is to train a driving agent who can execute `good' actions so that the new state \textit{and} possible state transitions until a finite expectation horizon will yield a high cumulative reward. The overall goal is quite straightforward for driving: not making collisions and reaching the destination should yield a good reward and vice versa. It must be noted that RL frameworks are not greedy unless $\gamma=0$. In other words, when an action is chosen, not only the immediate reward but the cumulative rewards of all the expected future state transitions are considered.  

Here we employ DQN \cite{mnih2015human} to solve the MDP problem described above. The main idea of DQN is to use neural networks to approximate the optimal action-value function $Q(s,a)$. This $Q$ function maps the state-action space to $\mathbb {R}$. $Q: S \times A \rightarrow {\mathbb {R}} $ while maximizing equation \ref{eqn_classification}. The problem comes down to approximiate or to learn this $Q$ function. The following loss function is used for Q-learning at iteration $i$.


\vspace{-0.6 cm}
\begin{multline}
\label{eqn_Q_function}
L_{i}(\theta) = \\ \mathbb{E}_{(s, a, r)}\left[ \left( r + \gamma \underset{a_{t+1}}{\textrm{max}} Q^{\theta_{i}^{-}}(s_{t+1}, a_{t+1})-Q^{\theta_{i}}(s_{t}, a_{t}) \right) ^{2} \right] 
\end{multline}

Where Q-Learning updates are applied on samples $(s, a, r) \sim U(D)$. $U(D)$ draws random samples from the data batch $D$. $\theta_{i}$ is the Q-network parameters and $\theta_{i}^{-}$ is the target network parameters at iteration $i$. Details of DQN can be found in \cite{mnih2015human}.

\subsection{Integrating path planning into model-free DRL frameworks}

The main contribution of this work is the integration of path planning into DRL frameworks. We achieve this by modifying the state-space with the addition of $d$. Also, the reward function is changed to include a new reward term $r_{w}$, which rewards being close to the nearest waypoint obtained from the model-based path planner, i.e. a small $d$. Utilizing waypoints to evaluate a DRL framework were suggested in a very recent work \cite{osinski2019simulation}, but their approach does not consider integrating the waypoint generator into the model.

The proposed reward function is as follows.  

\begin{equation}
\label{eqn_reward}
	r= \beta_{c} r_{c} + \beta_{v}r_{v} + \beta_{l}r_{l} + \beta_{w}r_{w}
\end{equation}

Where $r_{c}$ is the no-collision reward, $r_{v}$ is the not driving very slow reward, $r_{l}$ is being-close to the destination reward, and $r_{w}$ is the proposed being-close to the nearest waypoint reward. The distance to the nearest waypoint $d$ is shown in Figure \ref{fig_states}. The weights of these rewards, $\beta_{c}, \beta_{v} \beta_{l},\beta_{w},$ are parameters defining the relative importance of rewards. These parameters are determined heuristically. In the special case of $\beta_{c} = \beta_{v} = \beta_{l} = 0 $, the integrated model should mimic the model-based planner.

Please note that any planner, from the naive A* to more complicated algorithms with complete obstacle avoidance capabilities, can be integrated into this framework as long as they provide a waypoint. 

\section{Experiments}\label{sec_experiments}

As in all RL frameworks, the agent needs to interact with the environment and fail a lot to learn the desired policies. This makes training RL driving agents in real-world extremely challenging as failed attempts cannot be tolerated. As such, we focused only on simulations in this study. Real-world adaptation is outside of the scope of this work.

\begin{figure*}
	\centering
	\includegraphics[width=1.7\columnwidth]{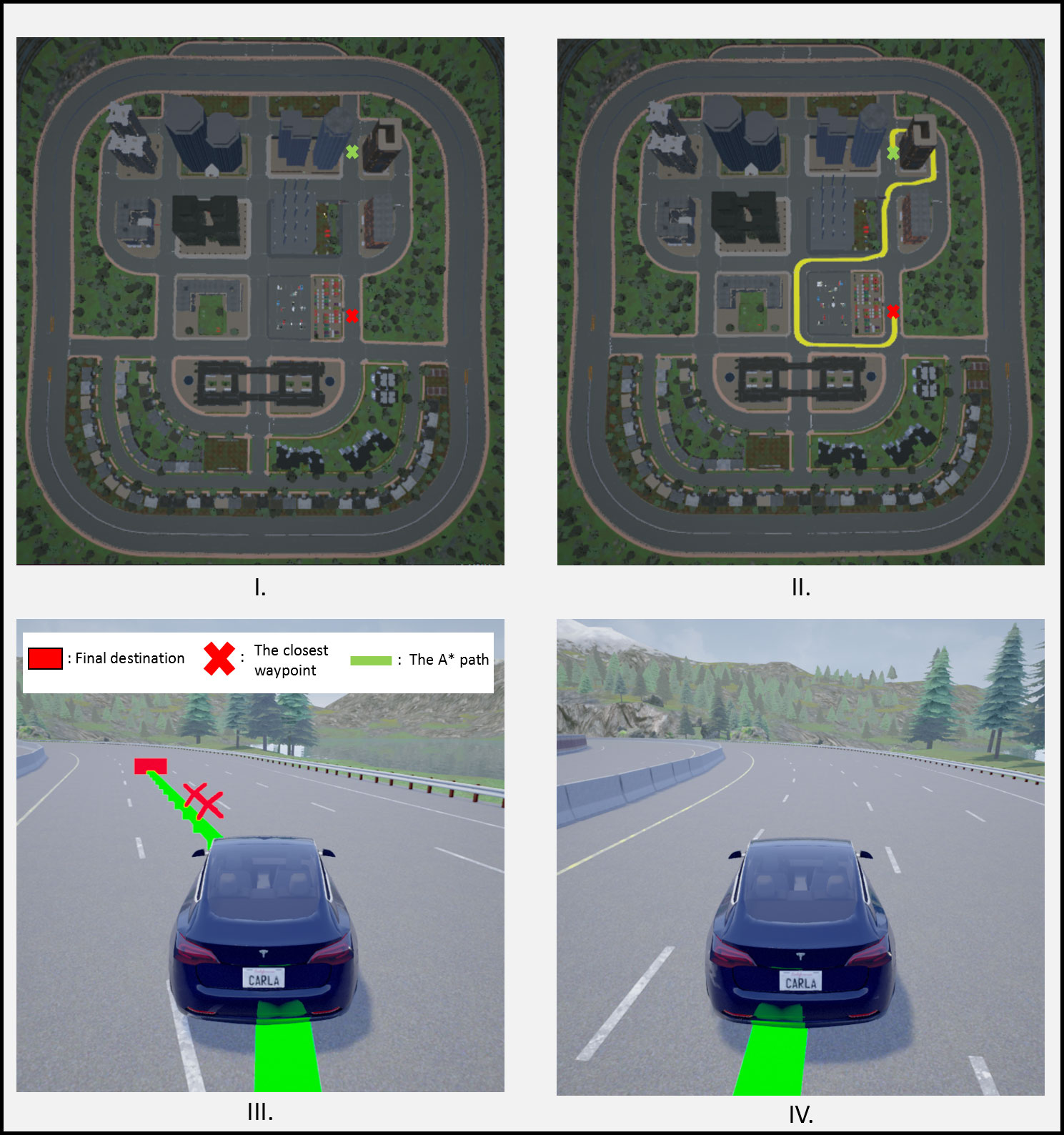}    
	\caption{The experimental process: I. A random origin-destination pair was selected. II. The A* algorithm was used to generate a path. III. The hybrid DRL agent starts to take action with the incoming state stream. IV. The end of the episode. }
	\label{fig_experiments}
\end{figure*}

The proposed method was implemented in Python based on an open-source RL framework \cite{SentdexGit} and CARLA \cite{dosovitskiy2017carla} was used as the simulation environment. The commonly used A* algorithm \cite{hart1968formal} was employed as the model-based path planner, and the recently proposed DQN \cite{mnih2015human} was chosen as the model-free DRL.

\subsection{Details of the reward function}

The general form of $r$ was given in the previous Section in equation \ref{eqn_reward}. Here, the special case and numerical values used throughout the experiments are explained.

\begin{equation}
r=\begin{cases} 
r_{v} + r_{l} + r_{w}& , r_{c} = 0 \textrm{ \& } l \geq \epsilon \\
100 &,  r_{c}=0 \textrm{ \& } l <\epsilon\\
r_{c} &, r_{c} \neq 0 
\end{cases}
\end{equation}

\begin{equation}
r_{c}=\begin{cases} 
0 & , \textrm{no collision}\\
-1 &, \textrm{there is a collsion} \\
\end{cases}
\end{equation}

\begin{equation}
r_{v} = \frac{1}{v_{0}}v-1
\end{equation}

\begin{equation}
r_{l}=1-\frac{l}{l_{\textrm{previous}}}
\end{equation}

\begin{equation}
r_{w} = 1-\frac{d}{d_{0}}
\end{equation}

Where $\epsilon=5m$, the desired speed $v_0=50 km/h$, and $d_{0}=8 m$. In summary,  $r_{w}$ rewards keeping a distance less than $d_{0}$  to the closest waypoint at every time step, and $r_{l}$ rewards decreasing $l$ over $l_{\textrm{previous}}$, distance to the destination in the the previous time step. The last term of $r_{l}$ allows to continuously penalize/reward the agent for getting further/closer to the final destination. 

If there is a collision, the episode is over and the reward gets a penalty equal to $-1$. If the vehicle reaches its destination $\exists \epsilon>0: l<\epsilon$, a reward of $100$ is sent back. Otherwise, the reward consists of the sum of the other terms.  $d_{0}$ was selected as $8m$ because the average distance between waypoints of the A* equals to this value.

\subsection{DQN architecture and hyperparameters}

The deep neural network architecture employed in the DQN is shown in Figure \ref{fig_dqn}. The CNN consisted of three identical convolutional layers with 64 filters and a $3\times3$ window. Each convolutional layer was followed by average pooling. After flattening, the output of the final convolutional layer, ego-vehicle speed and distance to the closest waypoint were concatenated and fed into a stack of two fully connected layers with 256 hidden units. All but the last layer had rectifier activation functions. The final layer had a linear activation function and outputed the predicted Q values, which were used to choose the optimum action by taking $\underset{Q}{\textrm{argmax}}$.

\subsection{Experimental process \& training}

The experimental process is shown in Figure \ref{fig_experiments}. The following steps were carried repeatedly until the agent learned to drive.

\begin{enumerate}
	\item Select two random points on the map as an origin-destination pair for each episode
	\item Use A* path planner to generate a path between origin-destination using the road topology graph of CARLA.
	\item Start feeding the stream of states, including distance to the closest waypoint, into the DRL agent. DRL agent starts to take actions at this point. If this is the first episode, initialize the DQN with random weights. 
	\item End the episode if a collision is detected, or the goal is reached.
	\item Update the weights of the DQN after each episode with the loss function given in equation \ref{eqn_Q_function}.
	\item Repeat the above steps sixty thousand times
\end{enumerate}

\subsection{Comparision and evaluation}

The proposed hybrid approach was compared against a complete end-to-end DQN agent. The complete end-to-end agent took only monocular camera images and ego-vehicle speed as input. The same network architecture was employed for both methods.  

A human driving experiment was also conducted to serve as a baseline.  The same reward function that was used to train the DRL agent was used as the evaluation metric. Four adults aging between 25 to 30 years old participated in the experiments. The participants drove a virtual car in CARLA using a keyboard and were told to follow the on-screen path (marked by a green line).  The participants did not see their scores. Every participant drove each of the seven predefined routes five times. The average cumulative reward of each route was accepted as the  ``average human score."

%

\begin{figure}
	\centering
	\includegraphics[width=1\columnwidth]{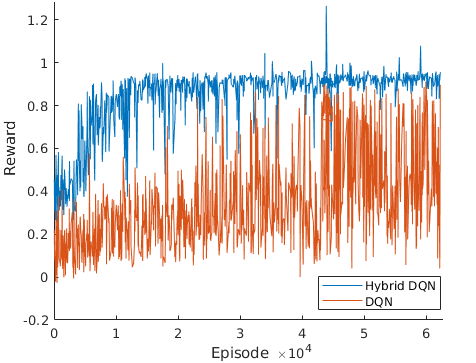}    
	\caption{Normalized reward versus episode number. The proposed hybrid approach learned to drive faster than its complete end-to-end counterpart.}
	\label{fig_result}
\end{figure}

\begin{table}[]
	\caption{Average reward scores for five runs in each route type.}
	\label{tab:reward}
	\centering
	\begin{tabular}{@{}l|c|c@{}}
		\multicolumn{1}{c|}{\textbf{Route type}} & \textbf{Hybrid-DQN} & \textbf{Human average} \\ 
		\hline
		Straight (highway) & 21.1  & 43.4  \\
		Straight (urban) &  27.6  & 38.1  \\
		Straight (under bridge)  & 31.6  & 45.2  \\
		Slight curve  & 30.4  & 49.5  \\
		Sharp curve  & -74.4  & -8.9  \\
		Right turn in intersection & -136.9  & -12.1  \\
		Left turn in intersection  & -385.9  & -25.5  \\ 
	\end{tabular}
\end{table}

\section{Results}\label{sec_results}

Figure \ref{fig_result} illustrates the training process. The result is clear and evident: The proposed hybrid approach learned to drive much faster than its complete end-to-end counterpart. It should be noted that the proposed approach made a quick jump at the beginning of the training. We believe the waypoints acted as a `guide' and made the algorithm learn faster that way. Our method can be used for spooling up the training process of a complete end-to-end variant with transfer learning. Qualitative analysis of the driving performance can be done by watching the simulation videos on our repository$^{\ref{footnote1}}$. 

The proposed method outperformed the end-to-end DQN, however, it is still not good as the average human driver as can be seen in Table \ref{tab:reward}.

Even though promising results were obtained, the experiments at this stage can only be considered as proof of concepts, rather than an exhaustive evaluation. The proposed method needs to consider other integration options, be compared against other state-of-the-art agents, and eventually should be deployed to the real-world and tested there.   

The model-based path planner tested here is also very naive. In addition, the obstacle avoidance capabilities of the proposed method was not evaluated. Future experiments should focus on this aspect. The integration of more complete path planners with full obstacle avoidance capabilities can yield better results.

\section{Conclusions}\label{sec_conclusion}

In this study, a novel hybrid approach for integrating path planning into model-free DRL frameworks was proposed. A proof-of-concept implementation and experiments in a virtual environment showed that the proposed method is capable of learning to drive. 

The proposed integration strategy is not limited to path planning. Potentially, the same state-space modification and reward strategy can be applied for integrating vehicle control and trajectory planning modules into model-free DRL agents.  

Finally, the current implementation was limited to output only discretized actions. Future work will focus on enabling continuous control and real-world testing.    


\section*{ACKNOWLEDGMENT}

This work was funded by the United States Department of Transportation
under award number 69A3551747111 for Mobility21: the National University
Transportation Center for Improving Mobility. 

Any findings, conclusions, or recommendations expressed herein are those
of the authors and do not necessarily reflect the views of the United
States Department of Transportation, Carnegie Mellon University, or
The Ohio State University.

%

\bibliographystyle{IEEEtran}
\bibliography{references_hybrid_deep_RL}

\end{document}